\documentclass[letterpaper, 10 pt, journal, twoside]{ieeetran}  

\usepackage{amsmath} 
\usepackage{amssymb}  
\usepackage{tikz}
\usepackage{graphicx}
\usepackage{lipsum}
\usepackage{caption}
\usepackage{subcaption}
\usepackage{color}
\usepackage{outlines}
\usepackage{siunitx}
\usepackage{bm}
\usepackage{multirow}
\usepackage{mathtools}
\usepackage{hyperref}
\usepackage{cite}
\usepackage{tabstackengine}
\setstackEOL{\cr}

\usepackage[font=footnotesize,labelfont=bf]{caption}
\makeatletter
\newcommand\footnoteref[1]{\protected@xdef\@thefnmark{\ref{#1}}\@footnotemark}
\makeatother

\title{\LARGE \bf
Optimal Initialization Strategies for Range-Only Trajectory Estimation
}

\author{Abhishek Goudar$^{1}$, Frederike D$\ddot{\textrm{u}}$mbgen$^{1}$, Timothy D. Barfoot$^{1}$, and Angela P. Schoellig$^{1,2}$
	\thanks{Manuscript received: September 15, 2023; Revised December 13, 2023; Accepted December 22, 2023.}%
	\thanks{This paper was recommended for publication by Editor Sven Behnke upon evaluation of the Associate Editor and Reviewers' comments.}
	\thanks{This work was supported in part by the Natural Sciences and Engineering Research Council of Canada (NSERC) and in part by the Canada CIFAR AI Chairs Program.}
	\thanks{$^{1}$ The authors are with the University of Toronto Institute for Aerospace 	Studies, Canada. They are also associated with the University of Toronto Robotics Institute and the Vector Institute for Artificial Intelligence. 
    \newline
    ${^2}$ The authors are with the Technical University of Munich, Germany. They are also associated  with the Munich Institute of Robotics and Machine Intelligence (MIRMI). E-mail: abhishek.goudar@robotics.utias.utoronto.ca,	\{frederike.dumbgen, tim.barfoot\}@utoronto.ca, angela.schoellig@tum.de.}%
	\thanks{Digital Object Identifier (DOI): see top of this page.}
}

\DeclareMathOperator*{\argmin}{arg\,min}
\newcommand{\trace}{\textrm{tr}}
\newcommand{\ltwonorm}{{L}_2}

\begin{document}

\maketitle

\begin{abstract}
Range-only (RO) pose estimation involves determining a robot’s pose over time by measuring the distance between multiple devices on the robot, known as tags, and devices installed in the environment, known as anchors. The non-convex nature of the range measurement model results in a cost function with possible local minima. In the absence of a good initial guess, commonly used iterative solvers can get stuck in these local minima resulting in poor trajectory estimation accuracy. In this work, we propose convex relaxations to the original non-convex problem based on semidefinite programs (SDPs). Specifically, we formulate computationally tractable SDP relaxations to obtain accurate initial pose and trajectory estimates for RO trajectory estimation under static and dynamic (i.e., constant-velocity motion) conditions. Through simulation and hardware experiments, we demonstrate that our proposed approaches estimate the initial pose and initial trajectories accurately compared to iterative local solvers. Additionally, the proposed relaxations recover global minima under moderate range measurement noise levels.  
\end{abstract}

\vspace*{-1em}
\section{Introduction}

\IEEEPARstart{R}{ange}-only (RO) localization involves determining the position of a mobile system, such as a robot, by measuring the distance between a range sensor on the robot, referred to as a \textit{tag}, and known landmarks in the environment, referred to as \textit{anchors}. Common examples of RO localization include radio frequency (RF)-based positioning such as the Global Positioning System (GPS) \cite{Kaplan2018} for outdoor environments, WiFi or ultrawideband (UWB)-based positioning for indoor environments \cite{indoorsurvey2019}, and acoustic positioning for underwater environments \cite{Papalia2022}. Since a single range measurement is not sufficient to estimate a robot's position and orientation, range sensors are typically combined with other sensing modalities such as wheel odometry \cite{Blanco}, inertial measurement units (IMUs) \cite{Hol2009}, and cameras \cite{Nguyen2021} to estimate the full pose. However, a common limitation of these sensor-fusion schemes is that sufficient motion is needed before the pose becomes observable \cite{Trawny2010, Goudar2021}. An alternative is to use multiple tags to estimate the full pose \cite{Goudar2023}, referred to as RO pose estimation. The advantage of such an approach is that motion is not necessary for pose estimation, but may still be beneficial. We use the term RO trajectory estimation to refer to both the estimation of the robot pose \textit{(i)}  at a single time step and \textit{(ii)} across multiple time steps while in motion, using only range measurements from multiple tags.

\begin{figure}[t]
	\centering
	\includegraphics[scale=1.4,trim={0cm 0cm 0 0.1cm}, clip]{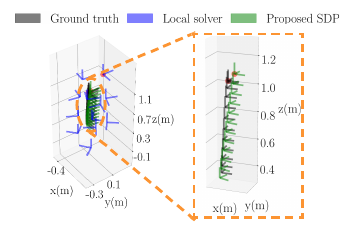}
	\vspace*{-1.4em}
	\caption{Range-only trajectory estimation results from a hardware experiment. A quadrotor in motion, equipped with 2 range sensors, measures its distance to multiple anchors to estimate its 3D position, velocity, and yaw angle over a short time horizon (pitch and roll come from an IMU). We refer to this as 2.5D dynamic initialization. The trajectory is parameterized by a sequence of poses, represented here by orthogonal axes. In the absence of a good initialization, the iterative local solver gets stuck in a local minimum resulting in poor accuracy compared to the ground-truth trajectory. Our proposed approach leverages a semidefinite relaxation of the original problem to recover accurate trajectories as shown in the magnified image on the right. For each method, the robot pose at $t=0$ is indicated by a red dot at the origin of the orthogonal axes.}
	\label{fig:dynamic_real_3d}
	\vspace{-1.6em}
\end{figure}

From a computational perspective, a common approach to RO localization is to formulate it as a maximum a posteriori (MAP) estimation problem, which results in the optimization of a particular objective function. In RO trajectory estimation, the non-convex nature of the range measurement model, along with an additive Gaussian measurement noise assumption, results in a non-convex nonlinear least-squares objective function that is typically optimized using local solvers such as the Gauss-Newton algorithm. A well-known limitation of such local solvers is the need for a good initialization point \cite{Trawny2010a, range_based_review}, without which the local solver can return suboptimal solutions, as shown in Figure \ref{fig:dynamic_real_3d}. 

The last decade has seen the development of algorithms that leverage Lagrangian duality theory~\cite{Boyd2013} to obtain \textit{convex} relaxations for non-convex problems found in many robotics applications. These convex relaxations typically take the form of semidefinite programs (SDPs) and can be used to obtain or certify globally optimal solutions. A tutorial on Lagrangian duality theory with a focus on robotics is provided in \cite{Barfoot2022}. 

In this work, we propose strategies to obtain accurate state estimates (including pose and velocities) for RO trajectory estimation based on semidefinite programming. Given the poor scalability of SDP solvers, we focus on computationally tractable SDP relaxations to obtain optimal initial poses and trajectories over a short time horizon. These initial estimates can be used to bootstrap local solvers that, given a good initial starting point, are generally accurate and more efficient over longer estimation horizons \cite{rosen_init_2015}. 

In contrast to previous work \cite{Dumbgen2022}, which focuses on position only and provides an optimality certificate for solutions from a local solver, our focus is to formulate tractable SDPs for \textit{pose} estimation that can be solved quickly for online applications. Additionally, unlike previous methods  \cite{Papalia2022, Papalia2023}, our method does not require additional odometry sensors; it can be used to generate initial poses without any motion, which we refer to as \textit{static initialization}. We also propose an initialization method for the case where a robot undergoes constant-velocity motion over a short time horizon; this is a challenging scenario for other methods that involve the fusion of a single range sensor and an odometry sensor as such motions lack the diversity that is required for observability \cite{Trawny2010}. We refer to initialization under such conditions as \textit{dynamic initialization}. 
To summarize, the following are the contributions of our work:
\begin{outline}
	\1 We present initialization strategies for range-only pose estimation under static conditions and range-only trajectory estimation under dynamic conditions involving constant-velocity trajectories. Our proposed initialization approaches recover optimal initial pose and trajectory estimates under moderate sensor-noise regimes. 
	\1 We validate the proposed methods in simulation and in multiple real experiments involving a ground robot and an aerial robot (see Figure \ref{fig:robots}).
\end{outline}

Qualitative experimental results of our proposed method, including its application as a mode of initialization for a fixed-lag smoother can be found in the accompanying video\footnote[1]{\label{supp_video}\url{http://tiny.cc/opt_ro_init_video}}.
%
\section{Related Work} \label{sec:related_work}

The use of range measurements for positioning is a well-studied problem due to its widespread application in popular technologies such as GPS \cite{Kaplan2018}. Other commonly used technologies for RO localization include sonar, lidar, WiFi, and UWB~\cite{indoorsurvey2019}. As mentioned previously, range sensors are typically combined with other sensing modalities such as wheel odometry \cite{Blanco}, IMU \cite{Hol2009, Goudar2021}, or visual odometry \cite{Nguyen2021}. A limitation of such sensor-fusion methods is the need for diverse motion for full-state observability~\cite{Trawny2010, Goudar2021}. More recently, approaches based on Gaussian-process regression~\cite{barfootgp2014} have been applied to continuous-time RO position~\cite{Dumbgen2022} and pose~\cite{Goudar2023} estimation. The benefit of these approaches is that no other sensing modalities are needed as the pose is observable (with multiple tags) under static conditions and with trajectories that lack diverse motion. 

For maximum-likelihood estimation, the non-convex nature of the range measurement model could result in a cost function with local minima. To recover the global minimum, a weighted least-squares (WLS) approach to planar pose estimation using range sensors aided by odometry is presented in~\cite{Trawny2010a}. A RO approach to planar pose estimation that converges to the maximum-likelihood estimate with minimum variance is presented in \cite{haodong2022}. A trilateration-based approach to RO relative pose estimation using multiple range sensors is shown in \cite{fishberg2022multi}.

Over the last decade, Lagrangian duality theory has been shown to be a powerful tool to generate optimal solutions and to certify the optimality of candidate solutions. It has been applied to various problems in robotics such as pose-graph optimization \cite{ Carlone2016}, simultaneous localization and mapping (SLAM) \cite{Carlone2015}, synchronization over the special Euclidean group \cite{Rosen2020}, extrinsic calibration between two egomotion sensors \cite{Giamou2019}, and outlier-robust estimation \cite{yang2023tipmi}. A common feature of these methods is that they reformulate the MAP estimation problem as a quadratically constrained quadratic program (QCQP), for which SDP relaxations exist. 

The application of duality theory to generate a certificate of optimality for RO localization was recently described in~\cite{Dumbgen2022}. Subsequently, an approach to certifiably correct range-aided SLAM with pose variables was demonstrated in \cite{Papalia2023}. Unlike previous works, which focus on certifying batch solutions \cite{Dumbgen2022} or requiring additional sensing modalities \cite{Trawny2010a,Papalia2022,Papalia2023}, we focus on computationally tractable SDP relaxations to obtain optimal initial pose and trajectories over a short time horizon using  multiple tags on the robot. 

Other works have studied the SDP relaxation of RO pose estimation \cite{Jiang2018, Jiang2019, Wu2022}. An important limitation of these approaches is that the SDP relaxations are not typically tight; the solutions are not rank 1 and require a subsequent refinement procedure. In this work, we empirically show that the lack of tightness stems from insufficient \textit{redundant} constraints. We leverage the approach of \cite{autosdp} to find the necessary redundant constraints to generate rank-1 solutions to RO trajectory estimation under static and dynamic conditions. 
\vspace*{-0.5em}

\section{Problem Statement} \label{sec:problem_statement}

The objective of our work is to estimate an accurate initial pose and trajectory, for static and dynamic conditions, respectively, using range measurements only.  We assume that the robot is equipped with multiple tags ($\geq 2$ for 2D and $\geq 3$ for 3D) and that the position of the tags is known with respect to the robot body frame, $\mathcal{F}_b$. 

We make the following additional assumptions for dynamic initialization. We assume that the robot moves with a constant velocity for a short time duration $t_v$ and that the range measurements arrive periodically every $\Delta t_r$ seconds during this period. The value of $t_v$ is chosen to ensure sufficient measurements given a particular range measurement rate. In all cases, we assume that a sufficient number of noncollinear anchors ($\geq 4$) are available and that their positions with respect to the world frame, $\mathcal{F}_\mathcal{W}$, are known. We assume that the range measurements are corrupted with moderate noise levels commonly encountered in practice.

\section{Methodology} \label{sec:methodology}


In this section, we develop SDP-based relaxations to
the (non-convex) problems of RO pose and trajectory estimation. Solving these non-convex problems locally using iterative solvers can result in sub-optimal solutions when local solvers get stuck in local minima. In contrast, the proposed relaxations are convex and can be solved to global optimality in polynomial time with off-the-shelf solvers. We find empirically that the relaxations are also often approximately tight -- meaning that a viable estimate to the original, non-convex optimization problem can be extracted from the SDP solution.

We introduce the notation that will be used in the rest of the paper. The variable $d \in \{2, 3\}$ is used to represent the pose dimension. Rotations are represented using elements of the special orthogonal group $SO(d) = \{\mathbf{R} \in \mathbb{R}^{d \times d}, \mathbf{R}^{T} \mathbf{R} = \mathbf{I}_d, \det(\mathbf{R}) = 1 \}$, where $\mathbf{I}_d$ is the identity matrix of dimension $d$, and $\det(\cdot)$ is the determinant operation.  We represent the robot pose $\mathbf{T}(t)$ at time $t$ with elements of the special Euclidean group:
\begin{equation}
\mathbf{T}(t) = \begin{bmatrix}
\mathbf{R}(t) & \mathbf{p}(t) \\
\mathbf{0} & 1
\end{bmatrix} \in SE(d),
\end{equation}
where the pose is parameterized by its position $\mathbf{p}(t) \in \mathbb{R}^d$ and rotation $\mathbf{R}(t) \in SO(d)$. Elements of the corresponding Lie algebra are denoted by $\bm{\varpi} \in \mathfrak{se}(d)$ \cite{Barfoot2023}. The \textit{homogenized} version of a vector $\mathbf{p} \in \mathbb{R}^d$ is represented by $\bar{\mathbf{p}} = \left[\mathbf{p}^T~ 1 \right]^T \in \mathbb{R}^{d+1}$. The trace of a matrix $\mathbf{A}$ is denoted by $\textrm{tr}(\mathbf{A})$. The Kronecker and the Hadamard products are denoted by $\otimes$ and $\odot$, respectively, and $\mathbf{X} \succcurlyeq 0$ implies $\mathbf{X}$ is a positive-semidefinite matrix.

\subsection{Range measurement model}
We assume $N_l > 1$ tags on the robot and seek to find its pose from measurements to $N_a$ anchors. The range measurement at any time $t$ is given by
\begin{equation}
 r_{jl}(t) = \| \mathbf{p}_{a_j} - \mathbf{K} \mathbf{T}(t) \bar{\mathbf{p}}_{u_l} \|_2, \nonumber
\end{equation}
where $\|\cdot\|_2$ is the $\ltwonorm$ norm, $\mathbf{T}(t)$ is the robot pose at time $t$, $\mathbf{p}_{a_j} \in \mathbb{R}^{d}$ is the position of anchor $j$ with respect to the world frame $\mathcal{F}_\mathcal{W}$, $\mathbf{p}_{u_l}$ is the position of tag $l$ with respect to the robot body frame $\mathcal{F}_b$, and $\bar{\mathbf{p}}_{u_l}$ is its homogeneous form. The matrix $\mathbf{K}$ is such that $\mathbf{p} = \mathbf{K} \bar{\mathbf{p}}$. 
%
Similar to \cite{Trawny2010a, Dumbgen2022}, we consider a squared-distance measurement model:
\begin{equation}
\tilde{r}_{jl}(t)^2 = \| \mathbf{p}_{a_j} - \mathbf{K} \mathbf{T}(t) \bar{\mathbf{p}}_{u_l} \|^2_2 + \eta_r(t), 
\end{equation}
where $ \eta_r(t) \sim \mathcal{N}(0, \sigma_r^2)$ is additive white Gaussian noise of variance $\sigma_r^2$. Next, we derive the MAP formulation and the corresponding SDP relaxation for static initialization. 

\subsection{Static initialization}

\subsubsection{Maximum a posteriori inference}
The state for static initialization is the robot pose at a single time step. The MAP estimate is the robot pose that minimizes the following objective function
\begin{align}
	\mathbf{T}(t)_{\rm{MAP}}^* = \argmin_{\mathbf{T}(t) \in SE(d)} \frac{1}{N_r} ~ \sum_{(j,l) \in \mathcal{E}}^{} \frac{e_{jl}(t)^2}{\sigma_r^{2}},\label{eqn:static_map}
\end{align}
where $\mathcal{E} \subseteq \{ (j,l)\,|\, j=1,...,N_a,~l=1,...,N_l\}$ is the index set for all measured anchor-tag pairs, $N_r = | \mathcal{E} |$ is the total number of range measurements, and 
\begin{align}
	e_{jl}(t) = \tilde{r}_{jl}(t)^2 - \| \mathbf{p}_{a_j} - \mathbf{K} \mathbf{T}(t) \bar{\mathbf{p}}_{u_l} \|^2_2.
	\label{eqn:error_term}
\end{align}
The MAP problem \eqref{eqn:static_map} can be solved using a local solver such as the Gauss-Newton or Levenberg-Marquardt algorithm \cite[Section 9.2.5]{Barfoot2023}. Static initialization across multiple time steps can be done in a similar manner by running multiple instances of \eqref{eqn:static_map}, one for each time step. As mentioned earlier, in the absence of a good initialization point, the local solver might return a suboptimal robot pose. 

We now derive an SDP relaxation to our original problem~\eqref{eqn:static_map} following the approach presented in \cite{Dumbgen2022}, but extending it to accommodate pose variables instead of only position variables. First, we reformulate our original problem \eqref{eqn:static_map} as a QCQP for which SDP relaxations exist. We begin by making the substitution $\tilde{\mathbf{p}}_{u_l}(t) = \mathbf{K} \mathbf{T} \bar{\mathbf{p}}_{u_l}$, which we refer to as the \textit{lever-arm} substitution, and expand the error term \eqref{eqn:error_term}:
\begin{align*}
	e_{jl}(t) &= \tilde{r}_{jl}(t)^2 - \| \mathbf{p}_{a_j} - \tilde{\mathbf{p}}_{u_l} \|^2_2, \\
	&= \underbrace{\tilde{r}_{jl}(t)^2 - \| \mathbf{p}_{a_j} \|^2_2}_{d_{jl}(t)}  - \underbrace{\| \tilde{\mathbf{p}}_{u_l}(t) \|^2_2}_{z_{u_l}} + 2  \mathbf{p}_{a_j}^T   \tilde{\mathbf{p}}_{u_l}(t), \\
	&= d_{jl}(t) + \left[ 2 \mathbf{p}_{a_j}^T ~ -1 \right] \begin{bmatrix}
		\tilde{\mathbf{p}}_{u_l}(t) \\ z_{u_l}
	\end{bmatrix}\!,
\end{align*}
where we have introduced a second substitution, ${ z_{u_l} = \|\tilde{\mathbf{p}}_{u_l}(t) \|_2^2}$.
With the new substitutions, the error term~\eqref{eqn:error_term} is linear and the cost function \eqref{eqn:static_map} quadratic in the unknown vector $\left[\tilde{\mathbf{p}}_{u_l}(t)^T \quad z_{u_l}\right]^T$. We define the vector of unknowns $\mathbf{x}_l =  {[\tilde{\mathbf{p}}_{u_1}(t)^T ~ z_{u_1} \, \hdots \, \tilde{\mathbf{p}}_{u_{N_l}}(t)^T ~ z_{u_{N_l}}]}^T \in \mathbb{R}^{(d+1) N_l}$. Stacking error terms corresponding to measurements from all anchor-tag pairs, we obtain the following optimization problem equivalent to \eqref{eqn:static_map}:
\begin{equation}
\begin{aligned}
\min_{\substack{\mathbf{x}_l, \mathbf{p}(t),\\\mathbf{R}(t)}}
  \quad & \frac{1}{\sigma_r^{2} N_r}\| \mathbf{w} \odot (\mathbf{d} + \mathbf{P}_a ~ \mathbf{x}_l) \|_2^2 \\
	\textrm{s.t.}
	 \quad & \tilde{\mathbf{p}}_{u_l}(t) = \mathbf{K}\mathbf{T}(t) \bar{\mathbf{p}}_{u_l}, \qquad  l=1,...,N_l,\\
	 \quad & \|\tilde{\mathbf{p}}_{u_l}(t) \|_2^2 = z_{u_l}, \qquad ~~\quad  l=1,...,N_l, \\
	 \quad & \mathbf{R}(t)^T  \mathbf{R}(t) = \mathbf{I}_d,\\
	 \quad & \det(\mathbf{R}(t)) = 1,
	\label{eqn:partial_qcqp}
\end{aligned}
\vspace*{-0.5em}
\end{equation}
where 
\begin{equation*}
\setstacktabbedgap{1.5pt}
\mathbf{w} \mathbin{=} \begin{bmatrix}
\delta_{11}\\
\vdots \\
\delta_{N_a N_l}
\end{bmatrix}\!,
	\mathbf{d} \mathbin{=} \begin{bmatrix}
	d_{11}(t)\\
	\vdots \\
	d_{N_a N_l}(t)
	\end{bmatrix}\!, 
	\mathbf{P}_a \mathbin{=}  ~ \mathbf{I}_{N_l} \otimes \begin{bmatrix}
	2 \mathbf{p}_{a_1}^T ~~ -1 \\
	  \vdots ~   \\
	2 \mathbf{p}_{a_{N_a}}^T ~ -1 \\
	\end{bmatrix}\!,
\end{equation*}
and 
\begin{equation*}
\delta_{jl} = \begin{dcases}
1 \quad \textrm{if} ~ (j,l) \in \mathcal{E},\\
0 \quad \textrm{otherwise.}
\end{dcases}
\end{equation*}
The orthogonality ${(\mathbf{R}(t)^T  \mathbf{R}(t) = \mathbf{I}_d)}$ and the determinant constraints (${\det(\mathbf{R}(t)) = 1}$) associated with the rotation matrix are included as explicit constraints; with this formulation the domain of optimization is now a vector space.

\subsubsection{SDP relaxation} \label{sec:sdp_relax_static}
We define our new state as $\mathbf{x}  = \left[ \mathbf{x}_l^T ~ \textrm{vec}(\mathbf{R}(t))^T  ~ \mathbf{p}(t)^T ~ h \right]^T$, where $\textrm{vec}(\cdot)$ converts the matrix $\mathbf{R}(t)$ into a vector by stacking its columns, and $h$ is a \textit{homogenization} variable. With the new state, \eqref{eqn:partial_qcqp} can be written as a QCQP:
\begin{equation}
\begin{aligned}
q^*  = \min_{\mathbf{x}} \quad & \mathbf{x}^T \mathbf{Q} \mathbf{x}\\
\textrm{s.t.}
\quad & \mathbf{x}^T \mathbf{A}_0 \mathbf{x} = 1,\\
\quad & \mathbf{x}^T \mathbf{A}_i \mathbf{x} = 0, \quad i=1,...,(d + 1)N_l+1,\\
\quad & \mathbf{x}^T \mathbf{B}_j \mathbf{x} = 0, \quad j=1,...,d(d-1).\\
\end{aligned} \label{eqn:qcqp}
\end{equation}
The relation between constraints in \eqref{eqn:partial_qcqp} and \eqref{eqn:qcqp} is as follows. The matrix $\mathbf{A}_0$ represents the homogenization constraint, $h^2 = 1$, matrices $\{\mathbf{A}_i \,|\, i=1,...,d\,N_l\}$ correspond to the lever-arm constraints,  $\{\mathbf{A}_i \,|\, i=d\,N_l+1,..., (d+1)\,N_l + 1\}$ encode $z_{u_l}= \|\tilde{\mathbf{p}}_{u_l}(t) \|_2^2$, and $\{\mathbf{B}_j \,|\, j=1,...,d(d-1)\}$ encode the orthogonality and the determinant constraints on $\mathbf{R}(t)$. Details on formulation of the orthogonality and the determinant constraints as quadratic constraints can be found in  \cite{Carlone2016, Giamou2019}, and in Appendix \ref{appn:quad_formu_rot}.
To obtain an SDP relaxation for \eqref{eqn:qcqp}, we make the substitution $\mathbf{X} = \mathbf{x}\mathbf{x}^T$. This substitution can be enforced with a convex positive semidefiniteness constraint $\mathbf{X} \succcurlyeq 0$ and a non-convex rank constraint $\textrm{rank}(\mathbf{X}) = 1$. We \textit{relax} the rank constraint to obtain the SDP relaxation:
\begin{equation}
\begin{aligned}
\mathit{p}^* = \min_{\mathbf{X}} \quad & \trace ( \mathbf{Q}^T \mathbf{X})\\
\textrm{s.t.}
\quad & \trace ( \mathbf{A}_0 \mathbf{X}) = 1, \\
\quad & \trace ( \mathbf{A}_i \mathbf{X}) = 0, \quad i=1,...,2(d+1)N_l,\\
\quad & \trace ( \mathbf{B}_j \mathbf{X}) = 0, \quad j=1,...,2(d-1),\\
\quad & \mathbf{X} \succcurlyeq 0.
\end{aligned}
\label{eqn:sdp}
\end{equation}
The SDP relaxation \eqref{eqn:sdp} provides a \textit{lower bound} to our original problem \eqref{eqn:static_map}.
%
If the solution $\mathbf{X}^*$ of \eqref{eqn:sdp}, is such that $\text{rank}(\mathbf{X}^*) = 1$, then the global minimum to our original problem \eqref{eqn:static_map} can be recovered using $\mathbf{X}^* = \mathbf{x}^*{\mathbf{x}^*}^T$. In this case, we say that the SDP relaxation is \textit{tight}. 

\begin{figure}[t]
	\hspace*{-0.4em}
	\includegraphics[scale=0.7, trim={0 0 0 0},clip]{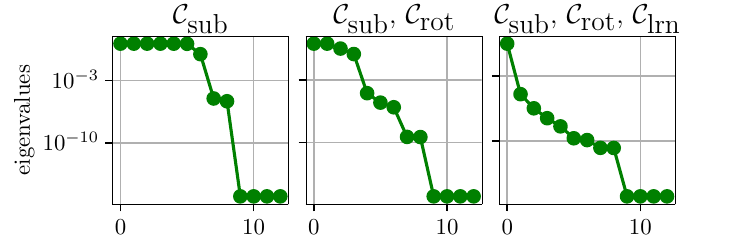}
	\caption{Qualitative results showing the effect of redundant constraints on the eigenvalue spectrum of the SDP solution. The set $\mathcal{C}_{\textrm{rot}}$ represents the orthogonality and the determinant constraints, and the set $\mathcal{C}_{\textrm{lrn}}$ denotes the additional redundant constraints \eqref{eqn:redundant_constraint}. In all cases, substitution constraints $\mathcal{C}_{\textrm{sub}}$ are included. The inclusion of additional redundant constraints results in a larger ratio of the dominant eigenvalues, and can lead to a rank-1 solution, as shown in the rightmost plot.}
	\label{fig:const_rank}
	\vspace{-1.5em}
\end{figure}

In general, the relaxation \eqref{eqn:sdp} may not be tight and we may need to incorporate additional \textit{redundant} constraints to obtain a rank-1 solution~\cite{teaser}. These constraints restrict the feasible set of our SDP relaxation \eqref{eqn:sdp} to favor rank-1 solutions, but do not affect the feasible set of the original problem \eqref{eqn:static_map}. We leverage the method of \cite{autosdp} to automatically generate the necessary redundant constraints by sampling the state space and identifying the nullspace associated with \eqref{eqn:sdp}. These constraints are added as additional constraints to the original SDP relaxation~\eqref{eqn:sdp} as
\begin{equation}
	\trace(\mathbf{S}_m \mathbf{X}) = 0, \qquad \forall m,
	\label{eqn:redundant_constraint}
\end{equation}
where each $\mathbf{S}_m$ encodes one redundant constraint. Examples of some redundant constraints for our method can be found in Appendix \ref{appn:redun_const}. Qualitative results from real 2D static initialization experiments showing the effect of the redundant constraints on the eigenvalue spectrum of the SDP solution are shown in Figure \ref{fig:const_rank}.

\subsection{Dynamic initialization} The method described so far requires the robot to be stationary to obtain an initial pose. This can be restrictive in certain dynamic settings such as initializing a quadrotor mid-flight. In dynamic initialization, the goal is to estimate the robot's trajectory, while in motion, over a short time horizon.

\subsubsection{Motion model} We use a constant-velocity motion model where the robot moves with a constant body-centric velocity for duration $t_v$. The motivation for such a motion model is to obtain SDPs that can be solved quickly for online applications. Specifically, under such a motion model, the robot trajectory is parameterized by the initial robot pose and the constant body-centric velocity. A similar approach has been adopted in \cite{Anderson2013}. Additionally, we assume range measurements arrive periodically every $\Delta t_r$ during the time window $t_v$. The value of $t_v$ depends on the measurement rate and is selected to obtain sufficiently many range measurements. For example, with measurement frequencies greater than $ 12\,\si{Hz}$, the value of $t_v$ is less than $ 1\,\si{\second}$, during which the constant-velocity assumption is often a good approximation.  

Let $\mathbf{T}(t_1)$ represent the unknown initial robot pose and $\bm{\varpi} \in \mathfrak{se}(d) = [\mathbf{v}^T ~ \bm{\omega}^T]^T$ the generalized constant velocity,
%
where $\mathbf{v}$ and $\bm{\omega}$ represent the body-centric linear and angular velocities, respectively. Under the constant-velocity assumption, the robot pose at any time $t_{k}$ is given by
\begin{align}
\mathbf{T}(t_k) &=  \mathbf{T}(t_1) \exp((c_k \bm{\varpi})^{\wedge}),
\label{eqn:pose_tk}
\end{align} 
where $c_k = (k-1)\Delta t_r$, $\exp (\cdot)$ maps an element of the associated Lie algebra back to the Lie group, and $\left( \cdot \right)^\wedge$ maps a vector to the corresponding skew-symmetric matrix. 
%
%
%
%
Next, we modify the MAP formulation from static initialization to incorporate the motion model for dynamic initialization.

\subsubsection{Maximum a posteriori inference}

The measured distance between anchor $j$ and the robot at time $t_k$ is
\begin{align}
\tilde{r}_{jl}(t_k)^2 &= \| \mathbf{p}_{a_j} - \mathbf{K} \mathbf{T}(t_k) \bar{\mathbf{p}}_{u_l} \|^2_2 + \eta_r(t)  \nonumber \\
&= \| \mathbf{p}_{a_j} - \mathbf{K} \mathbf{T}(t_1) \exp((c_k \bm{\varpi})^{\wedge}) \, \bar{\mathbf{p}}_{u_l} \|^2_2 + \eta_r(t), \nonumber
\end{align}
where the last line follows from \eqref{eqn:pose_tk}.
The corresponding error term for MAP inference is
\begin{equation}
e_{jl}(t_k) = \tilde{r}_{jl}(t_k)^2 - \| \mathbf{p}_{a_j} - \mathbf{K} \mathbf{T}(t_1) \exp((c_k \bm{\varpi})^{\wedge}) \, \bar{\mathbf{p}}_{u_l} \|_2^2. \nonumber
\end{equation}
%
%
%
The state for dynamic initialization is $\bm{\theta} = \{\mathbf{T}(t_1), \bm{\varpi}\}$ and the corresponding objective function for MAP inference is
\begin{equation}
\bm{\theta}_{\rm{MAP}}^* = \argmin_{\bm{\theta}} \frac{1}{N_r} \sum_{k=1}^{K} \sum_{(j,l)\in \mathcal{E}} \frac{e_{jl}(t_k)^2}{ \sigma_r^{2}}. 
\label{eqn:dynamic_map}
\end{equation}
\subsubsection{SDP relaxation} \label{sec:sdp_relax_dynamic}

A benefit of our constant-velocity motion model is that we can reuse the SDP relaxation we developed for static initialization, as the MAP objectives \eqref{eqn:static_map} and \eqref{eqn:dynamic_map} are similar. To account for motion across different time steps, the lever-arm constraints need to be modified to reflect the relationship between the initial pose, $\mathbf{T}(t_1)$, the body-centric velocity, $\bm{\varpi}$, and the pose at time $t_k$, $\mathbf{T}(t_k)$. Specifically, the new lever-arm constraint for time $t_k$ is
\begin{align}
\tilde{\mathbf{p}}_{u_l}(t_k) &= \mathbf{K}\mathbf{T}(t_k) \bar{\mathbf{p}}_{u_l} \nonumber \\
 &= \mathbf{K} \mathbf{T}(t_1) \exp((c_k \bm{\varpi})^{\wedge}) \, \bar{\mathbf{p}}_{u_l} \nonumber
 \\
&\approx \mathbf{K} \mathbf{T}(t_1) \left( \mathbf{I} + (c_k \bm{\varpi})^{\wedge}) \right) \, \bar{\mathbf{p}}_{u_l} \nonumber\\
&= \mathbf{K}  \mathbf{T}(t_1) \bar{\mathbf{p}}_{u_l} + c_k \mathbf{K} \mathbf{T}(t_1) (\bm{\varpi}^{\wedge}) \bar{\mathbf{p}}_{u_l} \label{eqn:lev_arm_const_approx},
\end{align}
where we have used a first-order approximation to the exponential map $\exp(\bm{\varpi}^\wedge) \approx {\mathbf{I}_d + \bm{\varpi}^\wedge}$. The above equation is quadratic in the unknowns and leads to an addition of $d\,N_r$ constraints to our SDP relaxation \eqref{eqn:sdp}. Note that with the first-order approximation, we are generating a lower bound to an approximation of our original problem \eqref{eqn:dynamic_map}; the motivation for doing so is to keep the computational complexity low. Details of an approximation-free approach can be found in Appendix~ \ref{appn:appr_free_dyn_init}.
\subsubsection{2.5D dynamic initialization} \label{sec:2.5d}
The size of the state for the SDP relaxation from Section \ref{sec:sdp_relax_dynamic} is larger compared to the one in Section \ref{sec:sdp_relax_static}. Furthermore, as in Section \ref{sec:sdp_relax_static}, we find that redundant constraints, identified using the tool described in \cite{autosdp}, are necessary for the solutions of the SDP relaxation of \eqref{eqn:dynamic_map} to be rank 1. The number of required redundant constraints grows quickly with the size of the state, which poses a computational challenge to the SDP solver.

To ensure a tractable initialization, we also consider 2.5D dynamic initialization where the robot pose is given by $\mathbf{T}(t) \in SO(2) \times \mathbb{R}^3$ and the corresponding body-centric velocity by $\bm{\varpi} = [\mathbf{v}^T ~ \bm{\omega}_z^T]^T$, with $\bm{\omega}_z ={[0 ~ 0 ~ \omega_z]}^T$. Specifically, only rotations around the body $z$-axis (yaw) and the corresponding angular rate are estimated. From a practical standpoint, this is adequate for most ground and aerial robots with built-in roll and pitch stabilization.

\section{Experiments} \label{sec:experiments}
In this section, we show the efficacy of our method through simulations and real experiments. In simulation, we show that our proposed initialization methods recover the global optimum under moderate range-measurement noise regimes. We then evaluate our approach on range data collected using multiple robots. 

To solve semidefinite programs, we use the CVXPY~\cite{cvxpy} package with the MOSEK~\cite{mosek} solver. As a baseline, we compare our method against MAP estimation with a custom implementation of the Levenberg-Marquardt solver. We compare our proposed approach (SDP) and the baseline local solver (LS) using $\ltwonorm$ position and rotation errors:
\begin{align}
	\text{Position Error} &= \|\mathbf{p}_{\rm{gt}} - \mathbf{p}_{\rm{est}} \|_2,\\
	\text{Rotation Error} &= \|\mathbf{R}_{\rm{gt}}^T \mathbf{R}_{\rm{est}}  - \mathbf{I}_d\|_F,
\end{align}
where $\mathbf{p}_{\rm{gt}}$ is the ground-truth position, $\mathbf{p}_{\rm{est}}$ is the estimated position, $\mathbf{R}_{\rm{gt}}$ and $\mathbf{R}_{\rm{est}}$ are the ground-truth and the estimated rotation matrices, and $ \| \mathbf{C} \|_F = \| \textrm{vec}(\mathbf{C}) \|_2$ is the Forbenius norm of matrix $\mathbf{C}$. As mentioned earlier, our SDP relaxation is tight if its solution has rank 1. In order to quantify the rank, we define the following ratio:
\begin{equation}
	f_{\rm{eig}}(\mathbf{X}) = \log_{10} \left(\frac{e_1}{e_2}\right),
	\label{eqn:eig_metric}
\end{equation}
where $e_1$ and $e_2$ are the dominant eigenvalues of the SDP solution $\mathbf{X}$, respectively. A large $f_{\rm{eig}}(\mathbf{X})$ suggests a rank-1 solution. All experiments are run on a laptop with an Intel Core i9 9750 CPU with 32 GB RAM.

\subsection{Simulation}
The objective of our simulation experiments is to demonstrate that our proposed approach recovers the global optimum while the baseline local solver is susceptible to local minima. Simulation parameters such as the measurement frequency and lever-arm configurations are selected to reflect real sensors and robots. Unless mentioned otherwise, we consider $N_l = 2$ tags in our experiments. The positions of the tags in the body frame are $\mathbf{p}_{u_1} = \left[0 ~ 0.095\right]^T\si{m}$ and $\mathbf{p}_{u_2} = \left[0 ~ -0.095\right]^T\si{m}$. 

\begin{figure}[t]
	\centering
	\begin{subfigure}[t]{0.5\textwidth}
		\centering
		\includegraphics[scale=0.85]{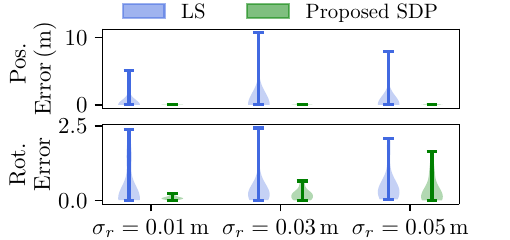}
		\caption{2D static initialization.}
		\label{fig:static_sim_2d}
	\end{subfigure}
	\begin{subfigure}[t]{0.5\textwidth}
		\centering
		\includegraphics[scale=0.85]{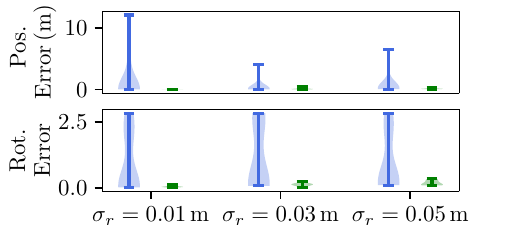}
		\caption{3D static initialization}
		\label{fig:static_sim_3d}
	\end{subfigure}
	\caption{Distribution of $\ltwonorm$ position and rotation errors from simulation for 2D static initialization (top) and 3D static initialization (bottom). The distributions are generated from 100 Monte Carlo trials across increasing range measurement noise, $\sigma_r$. The distribution of errors from the proposed method (SDP) is tighter compared to the baseline local solver (LS). The local solver accuracy is lower as it often gets stuck in local minima. The local solver typically converges to the global minimum when the robot is inside the convex hull of the anchors and performs poorly towards the boundary and outside the convex hull, whereas the proposed approach performs reliably even in such challenging scenarios.}
	\vspace*{-1.5em}
\end{figure}

\subsubsection*{Static initialization}
We perform multiple Monte Carlo simulations to evaluate the effectiveness of our proposed approach and the baseline in estimating an arbitrary initial pose. The setup for one trial is generated as follows. First, a random anchor configuration and a ground-truth robot pose are generated by \textit{(i)} sampling (robot and anchor) positions uniformly from $\left[-4, 4\right]\,\si{m}$ and \textit{(ii)} sampling robot rotations uniformly from $\left[-\pi, \pi \right]\,\si{rad}$. We then simulate range measurements between all possible anchor-tag pairs and corrupt the range measurements with Gaussian noise of increasing variance. For a given anchor configuration and ground-truth robot pose, we initialize the local solver with a random robot pose generated using the same procedure as before. Note that our proposed approach does not require an initial point.

For the 2D static initialization, we consider $N_a = 3$ anchors with $N_r = 6$ range measurements to the two tags, $\mathbf{p}_{u_1}$ and $\mathbf{p}_{u_2}$. Results from 100 simulation experiments are shown in Figure \ref{fig:static_sim_2d}. We see that the proposed method has a tighter distribution of errors compared to the local solver. The distribution of errors is larger for the local solver as it gets stuck in local minima. Additionally, solutions from our proposed method had $f_{\rm{eig}}(\mathbf{X}) \geq 7$ in all of our experiments, which we consider rank 1, indicating optimal solutions. In general, the local solver converges to the global minimum when the robot is inside the convex hull of the anchors and performs poorly outside the convex hull, whereas the proposed approach performs reliably even in such challenging scenarios.  
%

In 3D static initialization, we consider $N_a = 4$ anchors and $N_l = 3$ tags with $N_r = 12$ range measurements. The positions of the tags in the body frame are $\mathbf{p}_{u_1} = \left[0.01 ~ 0.41 ~ 0 \right]^T\si{m}$, $\mathbf{p}_{u_2} = \left[0 ~ -0.43 ~ 0.01\right]^T\si{m}$, and $\mathbf{p}_{u_3} = \left[-0.57 ~ 0.02 ~ 0\right]^T\si{m}$. Results from 100 simulations are shown in Figure~\ref{fig:static_sim_3d}. We see that the proposed method has a tighter distribution of errors compared to the baseline. Additionally, the ratio of the first two dominant eigenvalues of the SDP is large, $f_{\rm{eig}}(\mathbf{X}) \geq 8$, which we consider rank 1, indicating an optimal solution.

\begin{figure}[t]
	\vspace*{-1em}
	\centering
	\includegraphics[scale=0.85]{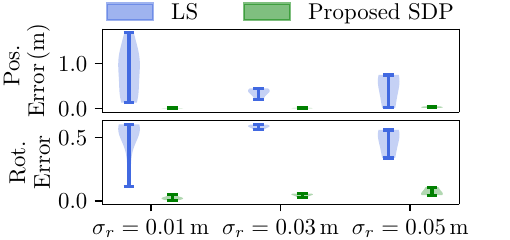}
	\caption{Distribution of $\ltwonorm$ position (top row) and rotation errors (bottom row) under increasing range measurement noise ($\sigma_r$) for 2D dynamic initialization from simulation. The position and rotation errors are computed over the full trajectory, which reflects any errors associated with the estimated velocity. The distribution of errors from the proposed method (SDP) is much tighter than the baseline local solver (LS) as the local solver gets stuck in local minima whereas the proposed method does not.} 
	\label{fig:dynamic_sim_2d}
	\vspace*{-0.5em}
\end{figure}

\begin{figure}[t]
	\hspace*{0em}
	\centering
	\includegraphics[scale=0.8, trim=0.0cm 0 0 0, clip]{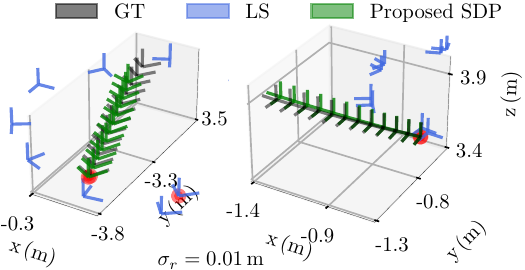}
	\caption{Two simulation results from 2.5D dynamic initialization showing the trajectories estimated by the iterative local solver (LS) and our proposed method (SDP) along with the ground-truth trajectory (GT). Without a good initial point, the local solver estimates suboptimal trajectories, while our proposed approach is able to generate better trajectory estimates. For each method, the robot pose at $t=0$ is indicated by a red dot at the origin of the orthogonal axes.}
	\label{fig:dynamic_sim_3d}
	\vspace*{-1.5em}
\end{figure}

\begin{figure}[t]
	\centering
	\includegraphics[scale=0.8, trim=0.65cm 0.2cm 0 0.0cm, clip]{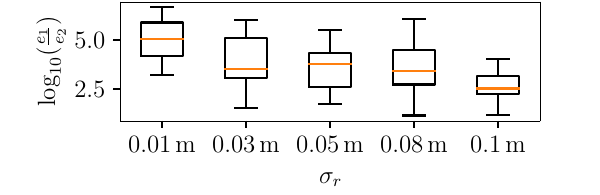}
	\caption{Simulation results from 2D static initialization quantifying the effect of measurement noise $\sigma_r$ on the optimality of the solution returned by the SDP solver. We quantify the optimality of an SDP solution as the logarithm of the ratio of its largest and second largest eigenvalues, denoted $e_1$ and $e_2$, respectively. As the magnitude of the measurement noise increases, the ratio of eigenvalues decreases.}
	\label{fig:eig_eval_2d}
	\vspace*{-1.5em}
\end{figure}

\subsubsection*{Dynamic initialization} The setup for a single trial of dynamic initialization is as follows. The anchor configuration and the initial robot pose are generated in a manner similar to the static case. In addition, a body-centric velocity is generated by uniformly sampling linear and angular velocities in the range $\left[-1,1\right]\,\si{\meter\per\si{\second}}$ and $\left[-0.3,0.3\right]\,\si{\radian\per\si{\second}}$, respectively. The initial pose and velocity are used to generate a constant-velocity trajectory for $t_v$ seconds with a pose every $\Delta t_r$ seconds. We then simulate range measurements between a tag and an anchor at each robot pose. Unlike the static case, there is a single range measurement at each time step.

The parameters for the 2D dynamic initialization are as follows. We consider $N_a = 3$ anchors, $N_l = 2$ tags, $N_r = 12$ range measurements, $t_v = 1.1\,\si{\second}$, and $\Delta t_r = 0.1\,\si{\second}$. The lever-arm configuration is the same as in the previous case. Results from 100 simulation experiments are shown in ~Figure \ref{fig:dynamic_sim_2d}. In this case, the position and rotation errors are calculated using all the poses in the robot trajectory, which captures any errors associated with the estimated velocity. Similarly to static initialization, the proposed method can estimate the trajectory reliably as indicated by a tighter distribution. The local solver is more susceptible to local minima in this case, potentially due to the sparse nature of the problem as there is a single range measurement at each time step. The distribution of dominant eigenvalue ratios is still relatively large, ${f_{\rm{eig}}(\mathbf{X}) \geq 5}$, with moderate measurement noise. We still consider it adequate to claim we have rank-1 solutions. 
 
\begin{table}[b]
	\vspace*{-1.em}
	\center
	\setlength\extrarowheight{2.0pt}
	\caption{Average computation time of the SDP optimizer and the local solver (LS) for static (stat) and dynamic (dyn) initialization in simulation.}
	\begin{tabular}{ | c | c | c | c | c |} 
		\hline
		& 2D stat. & 3D stat.  & 2D dyn.  & 2.5D dyn.  \\
		\hline
		SDP ($\si{\second}$) & $0.02$ & $0.05$ &  $0.83$ & $3.15$ \\
		\hline
		LS ($\si{\second}$) & $0.003$ & $0.006$ &  $1.4$ & $2.22$ \\
		\hline
	\end{tabular}
	\label{tab:avg_solv_time}
\end{table}
 
For 2.5D dynamic initialization, we consider $N_a = 4$ anchors and $N_l = 2$ tags. The rest of the parameters are identical to the previous case. Qualitative results showing estimated trajectories with $\sigma_r = 1\,\si{cm}$  from two such experiments are presented in Figure \ref{fig:dynamic_sim_3d}. The local solver is particularly susceptible to poor initializations in this case, while the proposed approach recovers the initial trajectory reliably. Additional result from simulation with $\sigma_r = 5\,\si{cm}$ is provided in Appendix \ref{appn:dyn_25d_eig_sim}.

The average time required by the SDP optimizer and the local solver for different initialization methods is presented in Table \ref{tab:avg_solv_time}. The time taken for 2.5D dynamic initialization is on the higher end due to the additional redundant constraints. However, 2.5D dynamic initialization can still be used to perform delayed initialization of local solvers where older states are initialized followed by an application of a forward motion model until the current time step.

\subsubsection*{Effect of noise on optimality} To further quantify the effect of noise on the optimality of the solution returned by the SDP solver, we performed multiple 2D dynamic initialization experiments with varying measurement noise magnitudes. Results from 100 experiments across 5 different measurement noise levels are shown in Figure \ref{fig:eig_eval_2d}. As the magnitude of measurement noise increases, the solution returned by the SDP solver is no longer obviously rank 1. 

\vspace*{-0.8em}

\subsection{Hardware experiments}
\begin{figure}[t]
	\centering
	\includegraphics[scale=0.148]{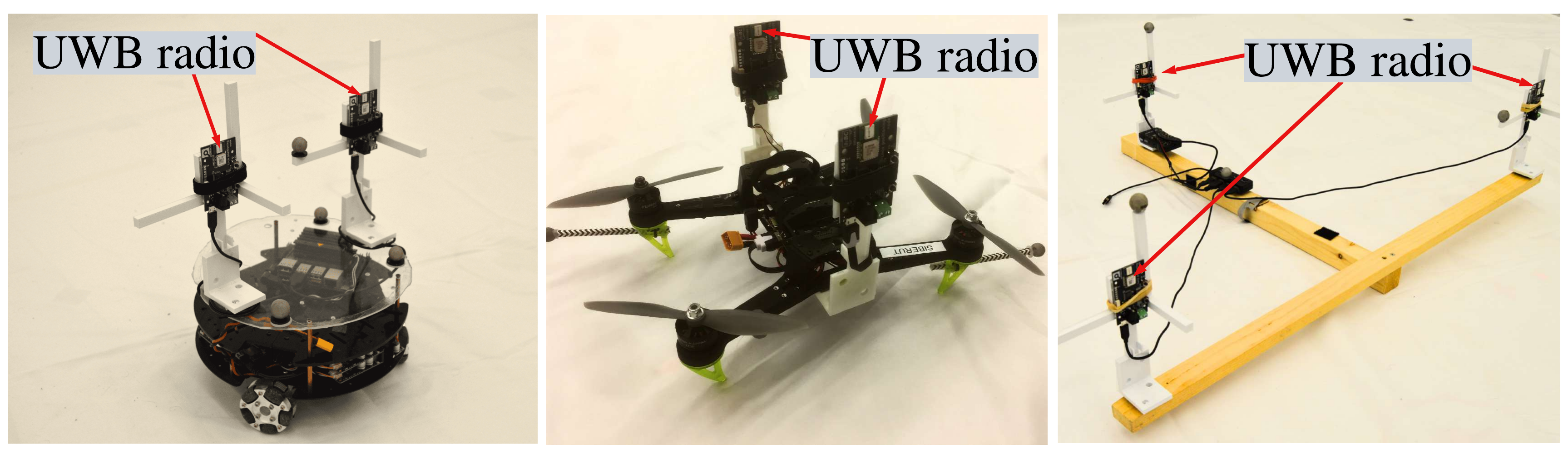}
	\caption{Our test platform for \textit{(i)} 2D static and dynamic initialization experiments is a ground robot with two ultrawideband (UWB) tags (left), \textit{(ii)} 2.5D dynamic initialization experiments is a quadrotor with two tags (centre), and \textit{(iii)} 3D static initialization experiments is a wand with three tags (right).}
	\label{fig:robots}
\end{figure}

Our test space is an indoor flight arena with 6 UWB anchors at the corners of a room of dimensions $7\,\si{m} \times 8\,\si{m} \times 3.5\,\si{m}$. The arena is equipped with a Vicon motion capture system for ground truth. We use the following test platforms: \textit{(i)} a ground robot with two tags for 2D static and dynamic initialization, \textit{(ii)} a quadrotor with two tags for 2.5D dynamic initialization, and \textit{(ii)} a sensor wand with three tags for 3D static initialization (see Figure \ref{fig:robots}). The positions of the tags with respect to the body frame are the same as in simulation. We remove any constant biases in the range data using ground-truth information.

\begin{figure}[t]
	\vspace*{-0.5em}
	\begin{subfigure}[t]{0.5\textwidth}
	\centering
	\includegraphics[scale=0.8, trim={0 0.0cm 0 0},clip]{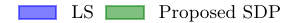}
	\end{subfigure}
	\begin{subfigure}[t]{0.25\textwidth}
		\includegraphics[scale=0.8, trim={0 0 0 0.2cm},clip]{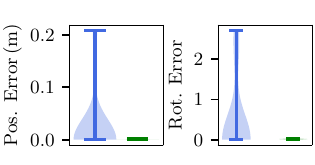}
		\vspace*{-1em}
		\caption{2D static initialization.}
		\label{fig:static_real_2d}
	\end{subfigure}%
	\begin{subfigure}[t]{0.25\textwidth}
		\includegraphics[scale=0.8,trim={0 0 0 0.2cm},clip]{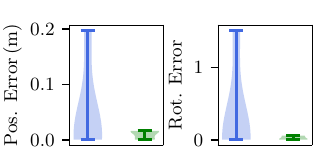}
		\vspace*{-1em}
		\caption{3D static initialization}
		\label{fig:static_real_3d}
	\end{subfigure}
	\caption{Distribution of $\ltwonorm$ position and rotation errors for 2D static initialization (left) and 3D static initialization (right) from real experiments. Error distributions for 2D static initialization are generated from 12 experiments where a ground robot's initial pose is estimated at different locations in the test space without any prior knowledge. For 3D static initialization, the distributions are generated from 10 experiments where a sensor wand's initial pose is estimated. The tighter spread of errors from the proposed method (SDP) shows that it recovers the initial pose consistently. The accuracy of the local solver (LS) is lower since it gets stuck in local minima.}
	\vspace*{-1.5em}
\end{figure}

%

\begin{figure}[t]
	\centering
	\vspace*{-0.4em}
	\hspace*{-1.5em}
	\includegraphics[scale=1.0]{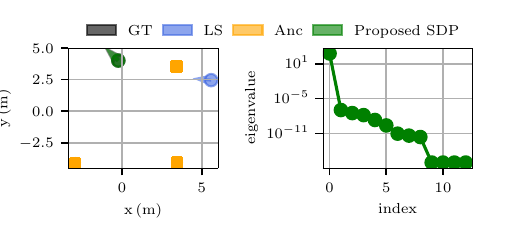}
	\vspace*{-1em}
	\caption{(Left) Estimated pose from real experiment for 2D static initialization. The pose estimated by the local solver (LS) and the proposed method (SDP) is shown along with the ground-truth pose (GT) and the anchor (Anc) positions. The GT pose overlaps with the SDP pose. (Right) The eigenvalue spectrum suggests a rank-1 solution, indicating a global minimum.}
	\label{fig:static_real_2d_pose}
	\vspace*{-1em}
\end{figure}

\subsubsection*{Static initialization}
For 2D static initialization, we performed multiple experiments by driving the ground robot to arbitrary positions in the test space. The range data and the ground-truth pose information were collected onboard for offline evaluation. The local solver is initialized with a random pose. Results from 12 experiments are shown in Figure \ref{fig:static_real_2d}. We see that the proposed method has a tighter distribution of errors compared to the baseline. The local solver performs poorly as it gets stuck in local minima. The solutions from our proposed method had $f_{\rm{eig}}(\mathbf{X}) \geq 7$, which we consider rank 1, indicating global minima. The estimated pose and the eigenvalue spectrum of the SDP solution from one such experiment are shown in Figure \ref{fig:static_real_2d_pose}. 
As in simulation, the local solver converges to the global minimum when the robot is inside the convex hull of the anchors and performs poorly outside the convex hull, whereas the proposed method performs reliably even in such conditions.

In 3D static initialization, we place the sensor wand at arbitrary poses at multiple locations in the test space and collect range data and ground-truth pose information for offline evaluation. Error plots from 10 experiments are shown in Figure \ref{fig:static_real_3d}. The proposed method has a tighter spread of errors compared to the local solver. Qualitative results along with the eigenvalue spectrum for 3D static initialization can be found in Appendix \ref{appn:real_3d_static}.


\subsubsection*{Dynamic initialization}

\begin{figure}[t]
	\centering
	\includegraphics[scale=0.8]{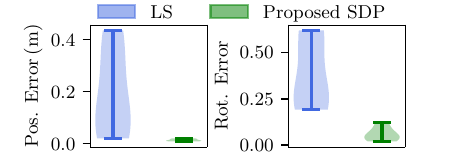}
	\caption{Distribution of $\ltwonorm$ position and rotation errors for 2D dynamic initialization from 5 real experiments. The narrow spread of errors for the proposed method (SDP) shows the efficacy of the proposed SDP relaxation in estimating the initial trajectory accurately compared to the local solver (LS), which gets stuck in local minima.}
	\label{fig:dynamic_real_2d}
	\vspace*{-1.5em}
\end{figure}

For 2D dynamic initialization, the ground robot is commanded along constant-velocity trajectories and the sensor data are recorded onboard for offline evaluation. The parameters for 2D dynamic initialization are the same as in simulation. Results from 5 real experiments are shown in Figure \ref{fig:dynamic_real_2d}. The proposed method estimates the robot trajectory accurately relative to the baseline. We observed that the measurement noise was higher than in simulation ($\sigma_r = 8\,\si{cm}$), leading to a drop in the ratio of the first two dominant eigenvalues with $f_{\rm{eig}}(\mathbf{X}) = 4$. However, even with a lower $f_{\rm{eig}}(\mathbf{X})$, the extracted solution candidates are superior to LS, as seen by the errors in Figure~\ref{fig:dynamic_real_2d}. Qualitative results including the eigenvalue spectra can be found in Appendix \ref{appn:2d_dyn_qual}.

\begin{table}[t]
	\centering
	\setlength\extrarowheight{2.0pt}
	\caption{Average $\ltwonorm$ position and rotation error of the  baseline local solver (LS) and the proposed method (SDP) from four hardware 2.5D dynamic initialization experiments.}
	\begin{tabular}{ | c | c | c |} 
		\hline
		{Algorithm} & {$\ltwonorm$ position error ($\si{m}$)} & {$\ltwonorm$ rotation error}\\
		\hline
		{LS} & 0.47 & 0.72\\
		\hline
		SDP & \textbf{0.02} & \textbf{0.07} \\
		\hline
	\end{tabular}
	\label{tab:dynamic_real_3d}
	\vspace*{-1em}
\end{table}

The parameters for 2.5D dynamic initialization are $N_a = 5$ anchors, $N_l = 2$ tags, $t_v = 1.1\,\si{\second}$, $\Delta t_r = 0.1\,\si{\second}$, and $N_r = 12$ range measurements. The tag positions in quadrotor body frame are $\mathbf{p}_{u_1} = \left[0 ~ 0.09\right]^T\,\si{m}$ and $\mathbf{p}_{u_2} = \left[0 ~ -0.09\right]^T\si{m}$. In each experiment, the quadrotor is commanded along constant-velocity trajectories and the sensor data and the ground-truth information are collected on the onboard computer. Ground-truth pose information from the Vicon system is used for closed-loop control of the quadrotor. We evaluated dynamic initialization on arbitrary segments of the trajectory. The local solver is initialized with the ground-truth pose in each case. Results from one such experiment is shown in Figure~\ref{fig:dynamic_real_3d}. The average position and rotation errors from four experiments are provided in Table \ref{tab:dynamic_real_3d}. As with 2D dynamic initialization, we observed that the measurement noise was high ($\sigma_r = 8\,\si{cm}$). The baseline LS performs poorly without a good initialization, while our method estimates the trajectories accurately. 

\vspace*{-0.8em}

\section{Conclusion and Future work} \label{sec:conclusion}

In this work, we presented approaches for estimating the initial pose and trajectory of a robot equipped with multiple range sensors. We developed semidefinite relaxations for range-only pose estimation under static conditions and range-only trajectory estimation for motion under constant-velocity trajectories. Through simulation and real experiments, we showed that the proposed relaxations achieve lower estimation error than local solvers and recover optimal initial poses and trajectories under moderate measurement noise. Because redundant constraints are required for tightness, our proposed methods are limited to low-dimensional states. Future work will look at trajectory estimation over longer horizons and attempt to use substitutions that require fewer redundant constraints. Another future direction involves exploring robust estimation approaches such as the truncated least-squares formulation~\cite{yang2023tipmi} to handle outlier range measurements.

\vspace*{-0.5em}

\bibliographystyle{unsrt}
\bibliography{main.bib}

\newpage
\appendix

\newcommand{\Rot}{\mathbf{R}}
\newcommand{\uvec}{\mathbf{r}_1}
\newcommand{\vvec}{\mathbf{r}_2}
\newcommand{\wvec}{\mathbf{r}_3}

This appendix is published along with the \textit{arXiv} verison of this paper and provides additional details and
results.

\subsection{Quadratic formulation of constraints of $SO(d)$ for $d=2$ and $d=3$} \label{appn:quad_formu_rot}

In this section, we provide details on formulating the constraints associated with the special orthogonal group $SO(d), d \in \{2, 3\}$ as quadratic constraints. Next, we show that nonconvex determinant constraint for $SO(3)$ is equivalent to the columns of the matrix satisfying the right-hand rule, which can be expressed as convex quadratic constraints. 

\subsubsection{Equivalence of the determinant and the right-handedness constraint for $SO(3)$} \label{appn:det_as_rh}

Consider a matrix $\Rot$ in the \textit{orthogonal} group:
\begin{align*}
O(3) = \{\Rot \in \mathbb{R}^{3} ~ | ~ \Rot^T \Rot = \mathbf{I}_3, \Rot \Rot^T = \mathbf{I}_3\}.
\end{align*}
We represent the matrix by its columns as $\mathbf{R} = \left[\mathbf{r}_1 ~ \mathbf{r}_2 ~ \mathbf{r}_3 \right]$,
%
%
where $\{\uvec, \vvec, \wvec \}\in \mathbf{R}^{3}$.
Then we have
\begin{align*}
\Rot^T \left( \uvec \times \vvec \right) &= \begin{bmatrix} \uvec^T \\ \vvec^T \\ \wvec^T \end{bmatrix} \left( \uvec \times \vvec \right) 
%
%
%
%
= \begin{bmatrix} 0  \\ 0 \\ \det(\Rot) \end{bmatrix},
\end{align*}
where $\times$ denotes the cross product and $\det(\cdot)$ denotes the determinant operator. The last equality follows from properties of scalar triple product. Multiplying by $\Rot$ on both sides, we get
\begin{align*}
\Rot ~ \Rot^T 	\left( \uvec \times \vvec \right) &= \det(\Rot) ~ \Rot \begin{bmatrix}
0 \\ 0 \\ 1
\end{bmatrix} \\
\uvec \times \vvec &= \det(\Rot) ~ \wvec.
\end{align*}
In a similar manner, we have
\begin{align*}
\wvec \times \uvec &= \det(\Rot) ~ \vvec,\\
\vvec \times \wvec &= \det(\Rot) ~ \uvec.
\end{align*}
From the above equations, the following equivalence holds:
%
\begin{equation}
\det(\Rot) =  1 \iff 
\begin{dcases}
\uvec \times \vvec = \wvec \\
\wvec \times \uvec = \vvec \\
\vvec \times \wvec = \uvec
\end{dcases} \label{eqn:det_as_rh}
\end{equation}\

\subsubsection{Quadratic formulation of orthogonality and right-handedness constraints }\label{appn:quad_form}

The determinant of any 2D matrix $\mathbf{R}$ is
\begin{equation*}
\det(\mathbf{R})	= r_1 r_4 - r_2 r_3,
\end{equation*}
where
\begin{equation*}
\mathbf{R} = \begin{bmatrix}
r_1 & r_2 \\
r_3 & r_4
\end{bmatrix}
\end{equation*}
Thus, the determinant constraint $(\det(\mathbf{R}) = 1)$ for a 2D rotation matrix can be expressed as a homogeneous quadratic equation:
\begin{align}
r_1 r_4 - r_2 r_3 - h^2 &= 0, \label{eqn:so2_det_const}
\end{align}
where $h$ is the homogenization variable with the constraint $h^2 = 1$. In this case, $\text{vec}(\mathbf{R}) =[r_1 ~r_3 ~r_2 ~r_4]^T$. For the state defined in static initialization, $\mathbf{x} = [ \mathbf{x}_l^T ~ \text{vec}(\mathbf{R}(t))^T ~ \mathbf{p}(t)^T ~ h ]$, the above equation can be represented as a quadratic constraint:
\begin{equation}
\mathbf{x}^T \mathbf{B}_j \mathbf{x} = 0,
\end{equation}
where the entries of the matrix $\mathbf{B}_j$ are the coefficients of the corresponding terms in \eqref{eqn:so2_det_const} with the rest of entries set to zero.

For a 3D rotation matrix, using the column representation from Appendix \ref{appn:det_as_rh} and the equivalence from \eqref{eqn:det_as_rh}, we can express the determinant constraint with the following homogeneous quadratic equations:
\begin{align*}
\left[ \mathbf{r}_i \right]_\times \mathbf{r}_j - \mathbf{r}_k h - \mathbf{1}h^2 &= 0, 
\end{align*}
where $\quad (i,j,k) \in \{(1, 2, 3), (3, 1, 2), (2, 3, 1)\}$,  $[\cdot]_\times$ maps a vector to the corresponding skew-symmetric matrix \cite{Barfoot2023} and $\mathbf{1}$ is a vector of ones of appropriate dimensions.

The orthogonality constraint ($\mathbf{R}^T \mathbf{R} = \mathbf{I}_d$) for both 2D and 3D rotation matrices can be written as the following homogeneous quadratic equations:
\begin{align}
\mathbf{r_i}^T \mathbf{r_i} - h^2 &= 0, \quad i \in \{1,...,d\},\\
\mathbf{r_i}^T \mathbf{r_j} &= 0, \quad i,j \in \{1,...,d\}, i \neq j,
\end{align}
where $d \in \{2,3\}$.

\subsection{Redundant constraints}\label{appn:redun_const}
We provide some examples of redundant constraints from our 2D static initialization setup. The lever-arm constraint can also be written as
\begin{equation}
\tilde{\mathbf{p}}_{u_l}(t) = \mathbf{R}(t) \mathbf{p}_{u_l} + \mathbf{p}(t),
\label{eqn:lev_arm_const_v2}
\end{equation}
We rewrite the above equation in expanded form, using the lever-arm configuration from our setup, as
\begin{align*}
\underbrace{
	\begin{bmatrix}
	\tilde{x}_{ul}\\
	\tilde{y}_{ul}
	\end{bmatrix}}_{\tilde{\mathbf{p}}_{ul}(t)} &=
\underbrace{
	\begin{bmatrix}
	r_1 & r_2 \\
	r_3 & r_4
	\end{bmatrix}}_{\mathbf{R}(t)} 
\underbrace{
	\begin{bmatrix}
	0\\
	y_{ul}
	\end{bmatrix}}_{ \mathbf{p}_{u_l}} 
+
\underbrace{
	\begin{bmatrix}
	x\\
	y
	\end{bmatrix}}_{\mathbf{p}(t)}.
\end{align*}
Examples of redundant constraints corresponding to \eqref{eqn:lev_arm_const_v2} are
\begin{align*}
(\tilde{x}_{ul} - {x})^2 &= (\tilde{x}_{ul} - x)(r_2 y_{ul}),\quad l=1,...,N_l,\\
(\tilde{y}_{ul} - {y})^2 &= (\tilde{y}_{ul} - y)(r_4 y_{ul}),\quad l=1,...,N_l.
\end{align*}
Some redundant constraints involving the rotation matrix are
\begin{align*}
r_{1} - r_{4} &= 0,\\
r_{2} + r_{3} &= 0,\\
z_{ul}(t)(r_{1} - r_{4}) &= 0.
\end{align*}
Using the homogenization variable, $h$, the above constraints and all others determined using the method of \cite{autosdp} can be expressed as quadratic constraints for inclusion in our SDP formulation.

\subsection{Approximation-free dynamic initialization}\label{appn:appr_free_dyn_init}
An exact SDP formulation for \eqref{eqn:dynamic_map}, without a first-order approximation to the exponential map  ${(\exp(\bm{\varpi}^\wedge) \approx {\mathbf{I}_d + \bm{\varpi}^\wedge})}$ can be obtained as follows. For each lever-arm substitution, $\tilde{\mathbf{p}}_{u} = \mathbf{K} \mathbf{T}(t_k) \bar{\mathbf{p}}_{u_l}$, we add the following additional constraints:

\begin{align}
\mathbf{T}(t_k) &= \mathbf{T}(t_{k-1}) \mathbf{\delta T}, \quad k=2\hdots K
\end{align}
where $\delta \mathbf{T} \coloneqq \exp((\bm{\varpi} \Delta t_r)^\wedge) \in SE(d)$. In addition, the orthogonality and the right-handedness constraints for each $\mathbf{R}(t_k)$ are included in our SDP formulation. The state is then extended to include $\{\mathbf{T}_k | k=1,...,N_r\}$ and $\delta \mathbf{T}$:
\begin{align*}
\mathbf{x}  = [ \mathbf{x}_l^T ~ \textrm{vec}(\mathbf{R}(t_1))^T  ~ \mathbf{p}(t_1)^T &\hdots \textrm{vec}(\mathbf{R}(t_K))^T  ~ \mathbf{p}(t_K)^T \\ &\textrm{vec}(\delta \mathbf{R})^T  ~ \delta \mathbf{p}^T ~ h ]^T 
\end{align*}
This formulation incurs higher computational cost due to the additional variables in the state. Specifically, we found that in the case of 2D dynamic initialization, the SDP optimizer took on average $12\,\si{\second}$ for the approximation-free approach, even without the inclusion of additional redundant constraints.


\subsection{Simulation}


\subsubsection{2.5D dynamic initialization}\label{appn:dyn_25d_eig_sim}

\begin{figure}[t]
	\centering
	\begin{subfigure}[t]{0.5\linewidth}
		\hspace*{0em}
		\includegraphics[scale=0.85, trim=0.cm 0 0 0, clip]{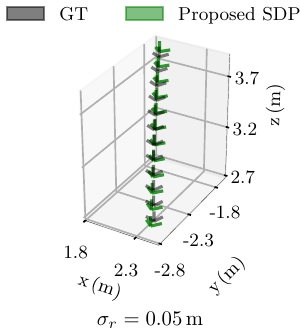}
	\end{subfigure}%
	\begin{subfigure}[t]{0.5\linewidth}
		\vspace*{-13em}
		\includegraphics[scale=0.85, trim=0cm 0 0 0, clip]{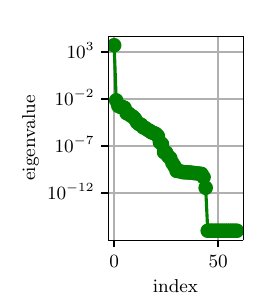}
	\end{subfigure}
	\caption{(Left) Simulation results from 2.5D dynamic initialization with range measurement noise $\sigma_r = 5\,\si{cm}$. The trajectory estimated by the proposed method (SDP) and the ground-truth trajectory (GT) are shown. The trajectory estimated by the local solver (LS) is not in the vicinity of the ground truth (GT) and hence is not visible. (Right) The eigenvalue spectrum of the SDP solution.}
	\label{fig:dynamic_sim_3d_eig}
\end{figure}

Qualitative results of the estimated trajectory and the eigenvalue spectrum with range measurement noise $\sigma_r = 5\,\si{cm}$ are shown in Figure \ref{fig:dynamic_sim_3d_eig}. The trajectory estimated by the SDP relaxation aligns with the ground-truth trajectory. The trajectory estimated by the local solver is not in the vicinity of the ground truth and hence is not visible.

\subsection{Real experiments}
\begin{figure}[t]
	\centering
	\includegraphics[scale=0.7]{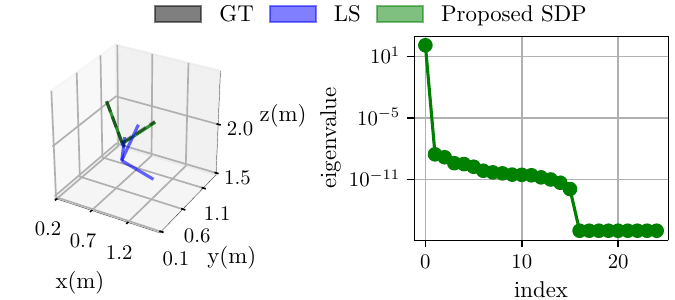}
	\caption{(Left) Qualitative results of the estimated pose from real 3D static initialization experiments. The pose estimated by the local solver (LS) and the proposed method (SDP) are shown along with the ground-truth pose (GT). The GT pose overlaps with the SDP pose. (Right) Log plot of the eigenvalue magnitudes of the corresponding SDP solution. The eigenvalue spectrum suggests a rank-1 solution, indicating that the proposed method recovers the global minimum.}
	\label{fig:static_real_3d_qual}
\end{figure}

\subsubsection{3D static initialization}\label{appn:real_3d_static}

Qualitative results from 3D static initialization showing the pose estimated by the local solver and the proposed method are shown in Figure \ref{fig:static_real_3d_qual}. The estimated pose  by the proposed method overlaps with the ground-truth pose. The eigenvalue spectrum of the SDP solution shown in Figure \ref{fig:static_real_3d_qual} suggests a rank-1 solution.


\begin{figure}[t]
	\centering
	\hspace*{-1.5em}
	\includegraphics[scale=0.6]{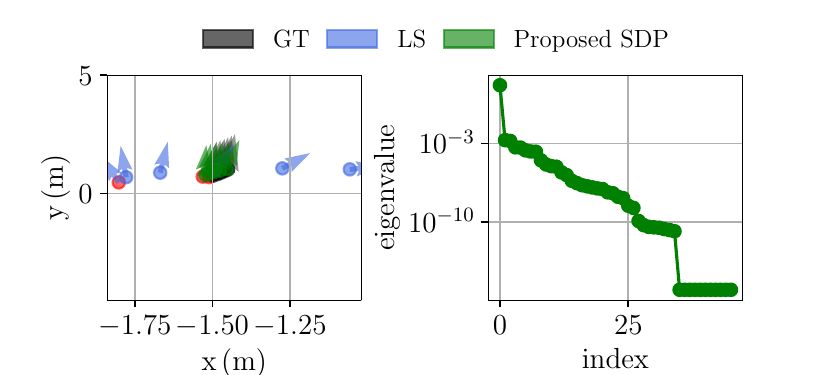}
	\caption{Results from a real 2D dynamic initialization experiment. (Left) Estimated trajectories from the local solver (LS) and the proposed method (SDP) along with the ground-truth trajectory (GT). The GT trajectory is not discernible as it overlaps with the SDP trajectory. For each method, the robot pose at $t=0$ is indicated by a red dot at the robot position. (Right) Log plot of the eigenvalue spectrum of the corresponding SDP solution.}
	\label{fig:dynamic_2d_result}
	\vspace*{-1.em}
\end{figure}

\subsubsection{2D dynamic initialization}\label{appn:2d_dyn_qual}

Results from 2D dynamic initialization with the trajectory estimated by the proposed method and the local solver are shown in Figure \ref{fig:dynamic_2d_result}. The eigenvalue spectrum of the SDP solution suggests a rank-1 solution.

\end{document}